\pdfoutput=1
\documentclass[pdflatex,sn-nature]{sn-jnl}

\usepackage{graphicx}%
\usepackage{multirow}%
\usepackage{amsmath,amssymb,amsfonts}%
\usepackage{amsthm}%
\usepackage{mathrsfs}%
\usepackage[title]{appendix}%
\usepackage{xcolor}%
\usepackage{tcolorbox}%
\usepackage{array}%
\usepackage{tabularx}%
\usepackage{textcomp}%
\usepackage{manyfoot}%
\usepackage{booktabs}%
\usepackage{hyperref}%
\usepackage{siunitx}%
\usepackage{caption}
\usepackage{float}
\newcounter{edfigure}
\newcounter{edtable}

\theoremstyle{thmstyleone}

\theoremstyle{thmstyletwo}

\theoremstyle{thmstylethree}

\begin{document}

\title[Learning before deployment]{A multi-domain physical AI agent for operating energy systems}

\author[1]{Yoon Pyo Lee}
\author[1]{Samrendra Roy}
\author[1]{Kazuma Kobayashi}
\author[1,3,4]{Souvik Chakraborty}
\author[1,2]{Syed Bahauddin Alam\textsuperscript{*}}\email{alams@illinois.edu}

\affil[1]{\orgdiv{Department of Nuclear, Plasma, and Radiological Engineering}, \orgname{The Grainger College of Engineering, University of Illinois Urbana-Champaign}, \orgaddress{\city{Urbana}, \state{IL}, \country{USA}}}
\affil[2]{\orgdiv{National Center for Supercomputing Applications}, \orgname{University of Illinois Urbana-Champaign}, \orgaddress{\city{Urbana}, \state{IL}, \country{USA}}}
\affil[3]{\orgdiv{Department of Applied Mechanics}, \orgname{Indian Institute of Technology Delhi}, \orgaddress{\city{New Delhi}, \country{India}}}
\affil[4]{\orgdiv{Yardi School of Artificial Intelligence}, \orgname{Indian Institute of Technology Delhi}, \orgaddress{\city{New Delhi}, \country{India}}}

\abstract{
Energy infrastructure is operated to achieve human objectives, but each objective must be expressed through physical commands specific to each system, and the operating knowledge that determines those commands is built separately for each system. Here we investigate a language-model-based multi-domain physical AI agent that learns from simulated and measured operating records to generate numerical commands together with physical explanations from human-readable states and objectives. One model without domain-specific modules operates across transient nuclear-reactor kinetics, Rankine-cycle turbine operation and a physical hydraulic-electric generator. It matches separately trained models, with mean absolute errors of 0.75~MW and 0.49~MW on reactor and turbine simulations and 0.62~mW on the first hardware request after each setpoint change. The same model controls the physical generator for 455 consecutive requests over 53.6~min with a tracking mean absolute error of 0.41~mW and no parameter updates, and sustained hardware operation is also achieved with a second language-model backbone. Each proposed command arrives with a learned physical rationale that an operator can review. Comparisons with the pretrained backbones show that joint training preserves the separation between domains while changing the representations within each domain. These results establish that operating capabilities governed by different physics can be acquired, explained and examined within one language model.
}

\maketitle

\section{Introduction}\label{sec:introduction}
Energy infrastructure is operated to achieve human objectives, but those objectives must ultimately be expressed through physical commands specific to each system. A requested change in power, for example, translates into reactivity insertion and its timing in a nuclear reactor, fuel and feedwater flows in a steam cycle, or pump actuation in a hydraulic-electric generator. Determining these commands requires operating knowledge that connects the requested outcome to the governing physics, available measurements and permissible actions. Established automation performs this translation through specialized control implementations, while human operators specify objectives, interpret system conditions and oversee execution. Artificial intelligence offers the prospect of bringing these distinct operating capabilities into a shared model, connecting human-readable states and objectives to quantitative physical actions through a common interface. This opportunity has acquired strategic urgency through government initiatives worldwide. The United States' Genesis Mission brings together national laboratories, industry and academia to accelerate AI-enabled scientific discovery and energy innovation, explicitly targeting advanced nuclear technologies, fusion and grid modernization \cite{doe2025genesis}. The United Kingdom's AI for Science Strategy identifies fusion energy as a priority, while the European Union's Apply AI Strategy connects AI adoption to the modernization of electricity infrastructure and management of energy assets \cite{uk2025aiforscience,ec2025applyai}. Translating this momentum into physical operation, however, introduces a consequential requirement. Once a command is executed, an incorrect action can interrupt electricity generation or compromise nuclear safety, and its effects cannot simply be discarded \cite{cnsc2024considerations,nrc2010regulatory}. Physical AI agents, which connect perception and decision-making to actions in the world \cite{physicalai2026editorial,durr2026outplaying}, must therefore acquire and demonstrate operating competence before consequential deployment \cite{iaea2021simulators,tang2025risks}. This motivates a shared operating intelligence that can accommodate different physical systems while preserving the separation between capability acquisition and operational execution.

Learned controllers now operate demanding physical systems. Deep reinforcement learning has controlled tokamak plasmas \cite{degrave2022magnetic} and fluid flows \cite{font2025deep,zolman2025sindy}, won champion-level drone races \cite{kaufmann2023champion} and won matches against elite table tennis players \cite{durr2026outplaying}. Learning from recorded operation has produced energy-management policies for fuel cell electric vehicles \cite{wang2025data}, plasma rampdown trajectories tested at the TCV tokamak \cite{wang2025learning} and real-time adjustment of battery cycling \cite{wang2025realtime}. In each case a controller is trained for one system and returns an action.

A shared operating intelligence would place the relationship between an objective and its physical implementation within a common model, while retaining the distinct requirements of each system. Language models offer this possibility, because system identity, measured state and requested outcome can be expressed through a common textual interface, and the model can generate the corresponding numerical command in a form that is both readable and executable. For a reactor, the required knowledge concerns the effect of reactivity and thermal feedback on power evolution. For a turbine, it concerns the relationship between fuel input, steam conditions and electrical output. For a hydraulic generator, it concerns how pump actuation changes the measured flow and generated power. Acquiring these relationships within shared parameters could allow one model to carry several operating capabilities without requiring a separate output architecture for each system. The same interface allows physical explanations to be learned together with the commands, so that each proposed action arrives with a stated physical rationale that an operator can review before the action is accepted. Government assessments, nuclear regulators and European law call for such human oversight of AI that acts on energy infrastructure and warn against over-reliance on its output \cite{doe2024airisk,cnsc2024considerations,euaiact2024}.

Recent advances have established several routes for connecting language-based intelligence to physical action. Embodied language models incorporate physical observations into language-based reasoning, while vision-language-action models translate visual observations and language instructions into executable robotic actions \cite{driess2023palm,brohan2023rt}. Language-model agents have operated an atomic force microscope \cite{mandal2025evaluating}, generated driving trajectories \cite{zhang2026autonomous} and controlled a manufacturing process \cite{margadji2026hybrid}. Their capabilities and limitations have been examined in the energy sector \cite{majumder2024exploring}, and our earlier work investigated a language-model-based physical AI agent whose operating strategies were evaluated through reactor simulation \cite{lee2026agentic}. Scientific agents generate hypotheses, coordinate investigations and revise proposals using the resulting evidence \cite{boiko2023autonomous,szymanski2023autonomous,gottweis2026accelerating,ghareeb2026multi}, while tool-using language agents incorporate observations from intermediate actions into subsequent decisions \cite{yao2022react,schick2023toolformer}. Table~\ref{tab:operating-scope} compares the tasks addressed in these studies and in the learned controllers above with those examined here. Together, these approaches connect language, decision-making and execution, and they leave three questions for energy-system operation. Can operating capabilities governed by different physics be acquired jointly within one language model, retain the accuracy of separately trained models and remain effective when executed repeatedly under physical feedback? Can physical explanations be learned together with the numerical actions they accompany? How are the acquired capabilities organized inside the model, and what does operational training change? Open weights permit the third question to be examined through probing classifiers and circuit-level analysis \cite{belinkov2022probing,elhage2021mathematical}.

Here we develop a multi-domain physical AI agent, a shared language-based operating intelligence learned from simulated and measured operating records spanning transient reactor kinetics, steady-state Rankine-cycle turbine operation and hydraulic-electric generation. A single dense model acquires the three operating mappings without domain-specific modules and retains the numerical accuracy of domain-specific models. We explicitly supervise physical explanations together with numerical actions conditioned on states and objectives, making the proposed command and its stated physical rationale available together for operator review. Comparisons with pretrained models show that joint training preserves pre-existing domain separation while changing representations within each domain. Accurate objective decoding for the reactor and turbine tasks coexists with different sensitivities of the original reactivity value's log-probability to reactor-target perturbations across backbones. Both OLMo and Granite sustain continuous operation of the physical generator, using new measurements to condition successive commands as their own actions change the measured state.

\begin{table}[tbp]
\centering
\caption{Operating tasks and scope of learned controllers and language-model agents for physical systems.}
\label{tab:operating-scope}
\footnotesize
\setlength{\tabcolsep}{4pt}
\begin{tabularx}{\linewidth}{@{}>{\raggedright\arraybackslash}p{0.20\linewidth}
  >{\raggedright\arraybackslash}X >{\raggedright\arraybackslash}X@{}}
\toprule
Study & Operating task addressed & Scope demonstrated \\
\midrule
\multicolumn{3}{@{}l}{\textit{Learned control of physical systems in Nature and Nature Communications}} \\
\addlinespace[2pt]
Degrave et al.~\cite{degrave2022magnetic} & Can deep reinforcement learning control a tokamak plasma? & A learned controller commands the full set of control coils and produces diverse plasma configurations on the TCV tokamak. \\
Kaufmann et al.~\cite{kaufmann2023champion} & Can a learned controller race a drone at champion level? & An autonomous drone wins several real-world races against three human champions. \\
D\"urr et al.~\cite{durr2026outplaying} & Can a robot play table tennis against elite players? & Model-free reinforcement learning with event-based vision wins matches against elite and professional players. \\
Wang, Y. et al.~\cite{wang2025data} & How should a fuel cell vehicle allocate power? & Offline reinforcement learning from recorded driving data, evaluated in powertrain simulation. \\
Wang, A. M. et al.~\cite{wang2025learning} & How should a tokamak plasma be ramped down? & Dynamics model learned from 311 recorded pulses, with trajectories tested at TCV. \\
Wang, M. et al.~\cite{wang2025realtime} & How should a battery be cycled to extend its life? & Closed-loop adjustment of charging on laboratory cells. \\
Font et al.~\cite{font2025deep} & Can deep reinforcement learning reduce a turbulent separation bubble? & Control learned on a coarse grid and applied on a fine grid in simulation. \\
Zolman et al.~\cite{zolman2025sindy} & Can learned control policies be made interpretable? & Compact sparse policies in five simulated environments. \\
\midrule
\multicolumn{3}{@{}l}{\textit{Language models acting on physical systems in Nature Communications}} \\
\addlinespace[2pt]
Mandal et al.~\cite{mandal2025evaluating} & Can language-model agents operate an atomic force microscope? & General-purpose agents operate the instrument through generated code. \\
Zhang et al.~\cite{zhang2026autonomous} & Can a language model plan driving trajectories? & Fine-tuned model evaluated on recorded data and in simulation. \\
Margadji and Pattinson~\cite{margadji2026hybrid} & Can a vision-language-action framework control a manufacturing process? & Language model with a regression network and retrieval, applied to a 3D printer. \\
\midrule
\textbf{This study} & Can one physical AI agent acquire, explain and execute the operating capabilities of energy systems governed by different physics? & One model matches separately trained models for a reactor, a turbine and a generator and explains each command. Two backbones each sustain 455 requests on the physical generator, and the domains remain separated inside the model. \\
\bottomrule
\end{tabularx}
\end{table}

\section{Results}\label{sec:results}
\subsection{A language interface separates learning from operation}\label{subsec:regimes}
\begin{figure}[H]
\centering
\includegraphics[width=\textwidth]{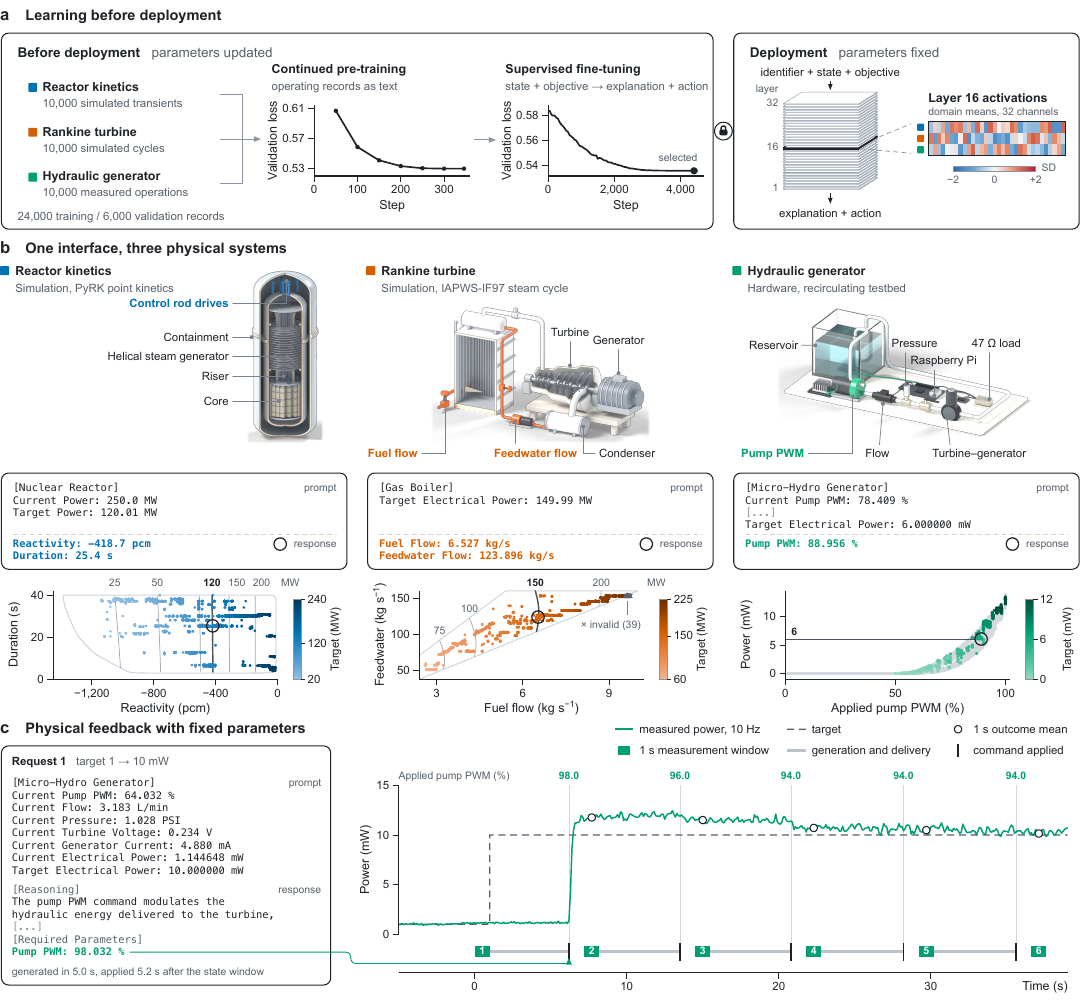}
\caption{\textbf{Learning before deployment across three physical systems.}
\textbf{a}, An OLMo-based AI agent is adapted by continued pre-training (CPT) on textual operating records and supervised fine-tuning (SFT) on explanations and actions conditioned on states and objectives, using 24,000 training and 6,000 validation records. Curves show recorded validation losses, whose magnitudes are not comparable between the two objectives. Points on the CPT curve mark evaluations, and the larger circle marks the selected SFT checkpoint. The padlock between the two frames marks deployment, after which parameters remain fixed. The read-out shows domain-averaged, standardized prompt representations of the joint model at decoder layer 16, with rows for reactor, turbine and generator from top to bottom and a shared scale from $-2$ to $+2$ channelwise standard deviations.
\textbf{b}, Device illustrations, verbatim prompts with response excerpts, and operating maps of the joint agent. Illustrations are schematic and not to scale, and the accent color marks the actuated components. Reactor and turbine maps show generated action pairs for 4,500 reactor requests and 1,250 turbine requests. Gray lines are iso-power contours of the forward response, interpolated from 10,000 executed PyRK records for the reactor and computed from the steady-state cycle model for the turbine, and are labeled in MW (Methods). Outlined regions mark the actions spanned by the reactor records and the thermodynamically valid turbine states, and crosses mark 39 commands outside the simulator's valid enthalpy range, all for targets above 218.4~MW, which are not part of the evaluation (Methods). The generator map shows measured electrical power versus applied pump pulse-width modulation (PWM) for 455 online requests during one 53.6~min run, over the 10,000 measured operations of the training corpus in gray. Point color denotes requested power in all three maps. Rings mark the requests quoted in the prompts above, bold values and dark lines give their requested power, and [...] marks omitted text.
\textbf{c}, Verbatim prompt and response excerpt for the first request of a 1-to-10~mW target transition in the same generator run, with the unsmoothed 10~Hz power trace for this and the four following requests. Numbers above the trace give applied PWM rounded to one decimal place, and open circles give 1~s outcome means. Each numbered 1~s measurement window serves as the outcome of one request and the state of the next. Decision intervals include generation and command delivery, and vertical marks denote command application. The target transition is drawn schematically at the end of the first state window, and time is measured from its start. Feedback updates the input state while model parameters remain fixed.}
\label{fig:regimes}
\end{figure}

Figure~\ref{fig:regimes} summarizes pre-deployment learning, operating mappings across three physical systems and feedback-driven execution with fixed model parameters. Continued pre-training on textual operating records is followed by supervised fine-tuning on explanations and actions conditioned on states and objectives (Fig.~\ref{fig:regimes}a) \cite{gururangan2020don}. At deployment, a textual domain identifier, the available state and the requested objective form a prompt. The model generates an explanation and a structured action. Each request uses one generation episode, with no simulator search, outcome-based revision of that action or parameter update.

The same interface accommodates reactivity and ramp duration for a reactor, fuel and feedwater flows for a turbine, and pump pulse-width modulation (PWM) for a hydraulic generator. Figure~\ref{fig:regimes}b shows verbatim prompts and responses, the joint agent's inferred reactor and turbine action pairs on iso-power contours of each simulator, and measured generator power versus applied pump PWM. The joint model uses a shared dense parameter space with no domain-specific adapters, output heads or expert modules. The textual identifier specifies the requested system. This formulation makes the input, proposed explanation and numerical command available for inspection while retaining a direct connection to physical execution.

\subsection{Three operating mappings coexist in one model}\label{subsec:three_systems}\label{subsec:joint}
\begin{figure}[H]
\centering
\includegraphics[width=\textwidth]{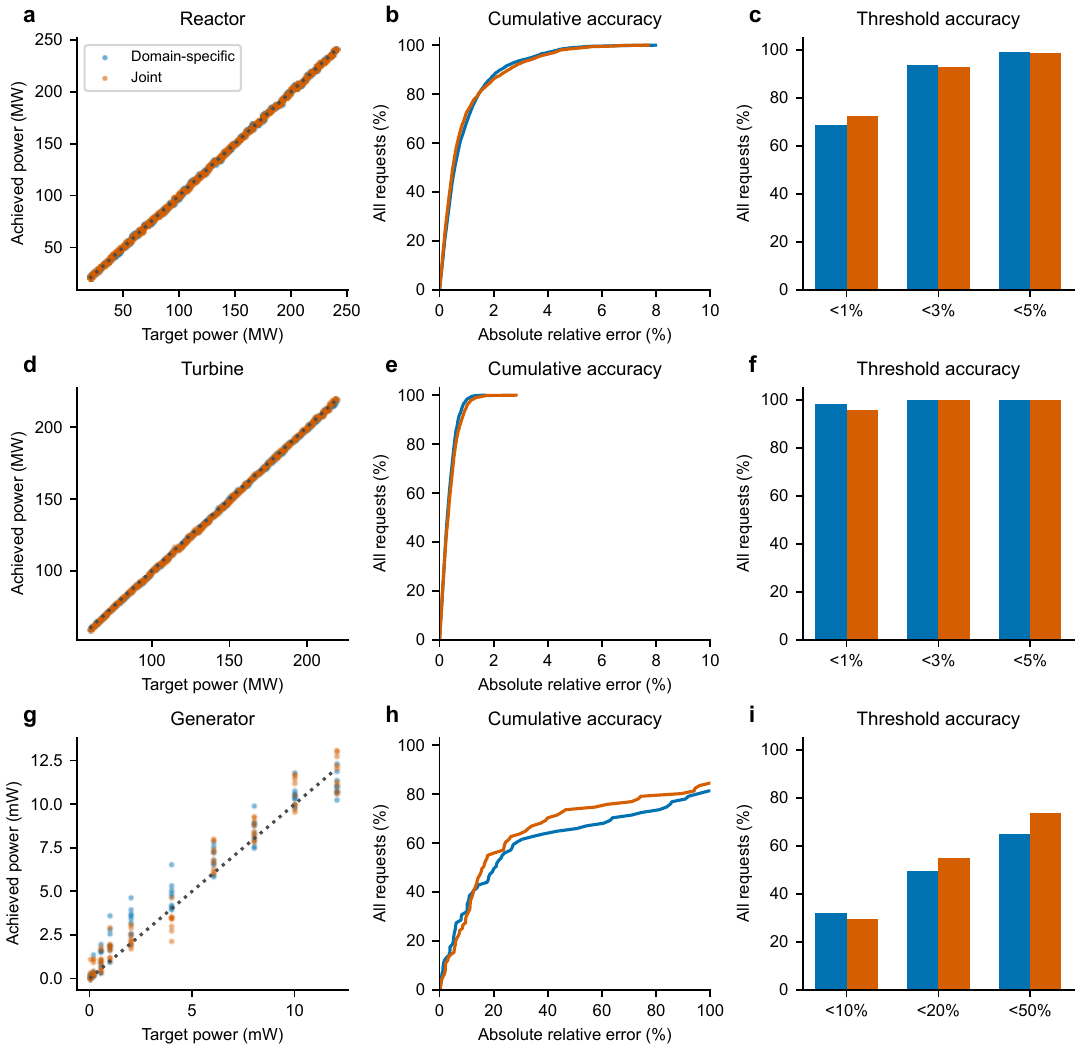}
\caption{\textbf{Domain-specific and joint inference across three physical systems.} Columns show target versus achieved power, cumulative accuracy and threshold accuracy. \textbf{a--c}, Reactor simulation ($n=4{,}500$ requests per model). \textbf{d--f}, Turbine simulation ($n=1{,}200$ requests per model). \textbf{g--i}, Generator measurements from the first request of each setpoint segment ($n=91$ per model). Identity lines are dotted. Curves and bars use all requests as their denominator. Hardware performance is evaluated online during the continuous operating sequence. Relative-error axes show 0--10\% for simulation and 0--100\% for hardware; requests outside these intervals remain in the denominators.}
\label{fig:three_systems}\label{fig:joint}
\end{figure}

The three systems examine whether operating mappings acquired before deployment can be executed across distinct physics, action variables and evaluation settings. We compare jointly trained and domain-specific models on the same requested target sequences to assess whether the acquired mappings remain usable within a shared parameter space, where transfer and interference may occur \cite{pmlr-v119-standley20a,NEURIPS2020_3fe78a8a}. The reactor task generated a reactivity insertion and ramp duration to reach a requested final power from an initial 250~MW state. We evaluated 4,500 targets spanning 20--240~MW, executing each action through a 300~s transient. The domain-specific model achieved a mean absolute error (MAE) of 0.84~MW and root mean squared error (RMSE) of 1.08~MW. The joint model achieved 0.75~MW and 0.93~MW, respectively (Fig.~\ref{fig:three_systems}a--c). The fractions within 1\% of target were 68.6\% and 72.5\%, and those within 5\% were 99.0\% and 98.6\%. All 4,500 trajectories from each model completed, with fuel and coolant temperatures below the specified evaluation limits.

The turbine task inferred fuel and feedwater mass-flow rates for 1,200 targets from 60 to 218.4~MW. Targets above this range lie at the edge of the simulator's valid property range, where almost no admissible action remains, and are not part of the evaluation (Methods). The domain-specific and joint models achieved MAE of 0.45~MW and 0.49~MW, with RMSE of 0.57~MW and 0.65~MW (Fig.~\ref{fig:three_systems}d--f).

The physical generator tested this interface with measured states and an actual actuator. The model received the current PWM, flow, pressure, voltage, current and power, together with a target electrical power, and generated a pump PWM command. Training used 10,000 recorded operations of the rig. Evaluation used a sequence over ten power levels from 0.05 to 12~mW. Single-request performance refers to the first request in each of 91 setpoint segments. The domain-specific model achieved MAE of 0.74~mW and median absolute relative error of 21.0\%; the joint model achieved 0.62~mW and 16.4\% (Fig.~\ref{fig:three_systems}g--i). Supplementary Fig.~1 resolves these errors by setpoint transition.

Together, these results demonstrate multi-domain operation within a single jointly trained model, with errors matching those of separately trained domain-specific models on identical requested target sequences.

\subsection{Commands are accompanied by reviewable physical rationales}\label{subsec:rationales}
Every response contains a physical explanation followed by the command (Fig.~\ref{fig:regimes}b,c), so the stated rationale is available before the command is executed. We examined the explanations generated by the joint OLMo model in the evaluations above. In the turbine and generator tasks the explanation restates the action it proposes. These values agreed with the executed command in all 1,200 turbine requests and in all 455 generator requests. The explanations also state physical quantities that can be checked. Reactor responses give the expected final fuel temperature and minimum coolant temperature of the transient. Against the PyRK execution of the same command, these had MAE of 0.39~$^\circ$C (maximum 1.76~$^\circ$C) and 0.08~$^\circ$C (maximum 0.33~$^\circ$C) over 4,500 requests. Turbine explanations give the thermal input delivered to the steam, which agreed with the value implied by the commanded fuel flow to within 0.10\% on average (maximum 0.75\%). Generator explanations give the turbine voltage and generator current associated with the requested power in 326 of the 455 requests. The values measured after the command was applied differed from them by MAE of 0.018~V and 0.62~mA. The explanations follow the form of their training annotations. Generator explanations word the stated voltage and current as measured values although they are written before actuation, and they rarely refer to the current measured state. Reactor explanations name the same physical mechanisms across requests, and their closing assessment line is identical in all 4,500 responses, as it is in the training records. Each proposed command therefore arrives with a stated physical rationale whose quantities an operator can check against independent information.

\subsection{Retained domain structure and differing objective sensitivity}\label{subsec:segregation}
The open-weight models allow the organization of the shared representation space to be examined alongside action performance. Studies of networks trained on multiple cognitive tasks have shown that task relationships can be reflected in the organization of their internal representations \cite{yang2019task}. We pooled prompt-token representations at the embedding output and every decoder output, then calculated silhouette scores after principal-component reduction, using 450 prompts from each domain. The pretrained backbone was evaluated on the same prompts as a reference for the structure present before operational training.

\begin{figure}[H]
\centering
\includegraphics[width=\textwidth]{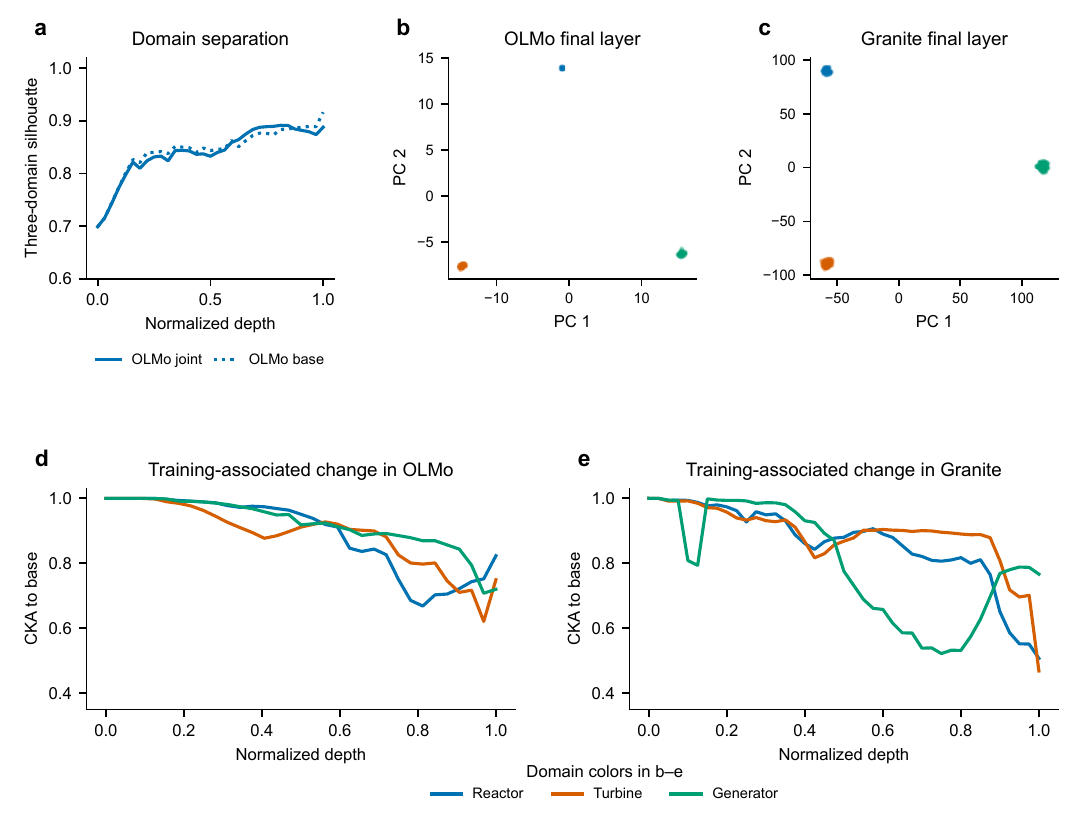}
\caption{\textbf{Domain structure persists while within-domain representations change.} \textbf{a}, Silhouette of mean-pooled prompt representations after PCA to at most 50 components, across normalized depth. The solid line denotes joint OLMo and the dotted line its pretrained backbone on identical prompts ($n=450$ per domain). \textbf{b,c}, Two-component PCA projections of the final-layer prompt representations for OLMo and Granite, respectively; each point is one prompt. PCA is fitted separately for each model, so axes are not shared between models. \textbf{d,e}, Linear CKA between pretrained and adapted representations, computed separately within each domain for OLMo and Granite. Depth zero is the embedding output and depth one is the final decoder output.}
\label{fig:segregation}
\end{figure}

For OLMo, the three-domain silhouette rose from 0.70 at the embedding output to 0.89 at the final layer (Fig.~\ref{fig:segregation}a). The pretrained backbone already separated these prompts, with final-layer silhouette 0.92. Joint training therefore preserved domain separability. Final-layer pairwise silhouettes were 0.90 for the reactor and turbine pair and 0.89 for each pair involving the generator. The final-layer projections visualize these domain-associated clusters (Fig.~\ref{fig:segregation}b,c).

Training also changed the geometry within each domain. Linear centered kernel alignment (CKA) between the pretrained and joint models reached final-layer values of 0.82, 0.75 and 0.72 for reactor, turbine and generator prompts, respectively (Fig.~\ref{fig:segregation}d). The larger Granite backbone retained domain separation with a final silhouette of 0.80, compared with 0.88 before adaptation; its corresponding within-domain CKA values were 0.51, 0.47 and 0.77 (Fig.~\ref{fig:segregation}e). A linear probe recovered reactor versus turbine domain identity with balanced accuracy 1.00 at every depth, and ridge probes recovered the objective with held-out $R^2>0.99$ from normalized depth 0.6 onward in joint OLMo, joint Granite and a joint OLMo variant trained by supervised fine-tuning without continued pre-training (Supplementary Fig.~3a,b).

Objective perturbations then assessed how the probability assigned to the original reactivity value changed when the reactor objective was altered. Masking the objective produced median log-probability changes of $-1.78$ for joint OLMo, $-2.98$ for the fine-tuning-only OLMo variant and $-1.49$ for Granite. For the two OLMo variants, a 1\% shift of the objective in either direction produced median changes of about $-0.64$ and $-0.44$, respectively, smaller in magnitude than their masking effects. For Granite, a 1\% shift produced a median change of $-1.48$, nearly equal to masking (Supplementary Fig.~3c). Thus, similarly accurate objective decoding coexisted with different sensitivities of the original reactivity value's probability to small objective changes. The sensitivity patterns differed between the two backbones and were qualitatively similar across the two OLMo training variants. Together with the pretrained comparisons, these observations distinguish retained domain structure, changes within that structure and differences in the objective dependence of numerical output probabilities.

\subsection{Fixed models sustain physical feedback control}\label{subsec:closed_loop}
Continuous operation examines whether a mapping acquired before deployment remains usable as the agent's own actions change the state encountered by subsequent requests. Each request applies the acquired mapping to a new measurement and objective, without deployment-time parameter learning. A continuous 10~Hz sensor stream supplies the latest 1~s state window. The frozen model generates a pump command, which is applied immediately; the outcome is measured in a 1~s window ending 2~s after application. Each of 91 target segments contains five successive requests, giving 455 requests per run. The model receives the updated state each time, but its parameters remain unchanged. Figure~\ref{fig:regimes}c shows the verbatim first request and the measurement and command timing for five successive requests during a selected 1-to-10~mW target transition in the joint OLMo run.

\begin{figure}[H]
\centering
\includegraphics[width=\textwidth]{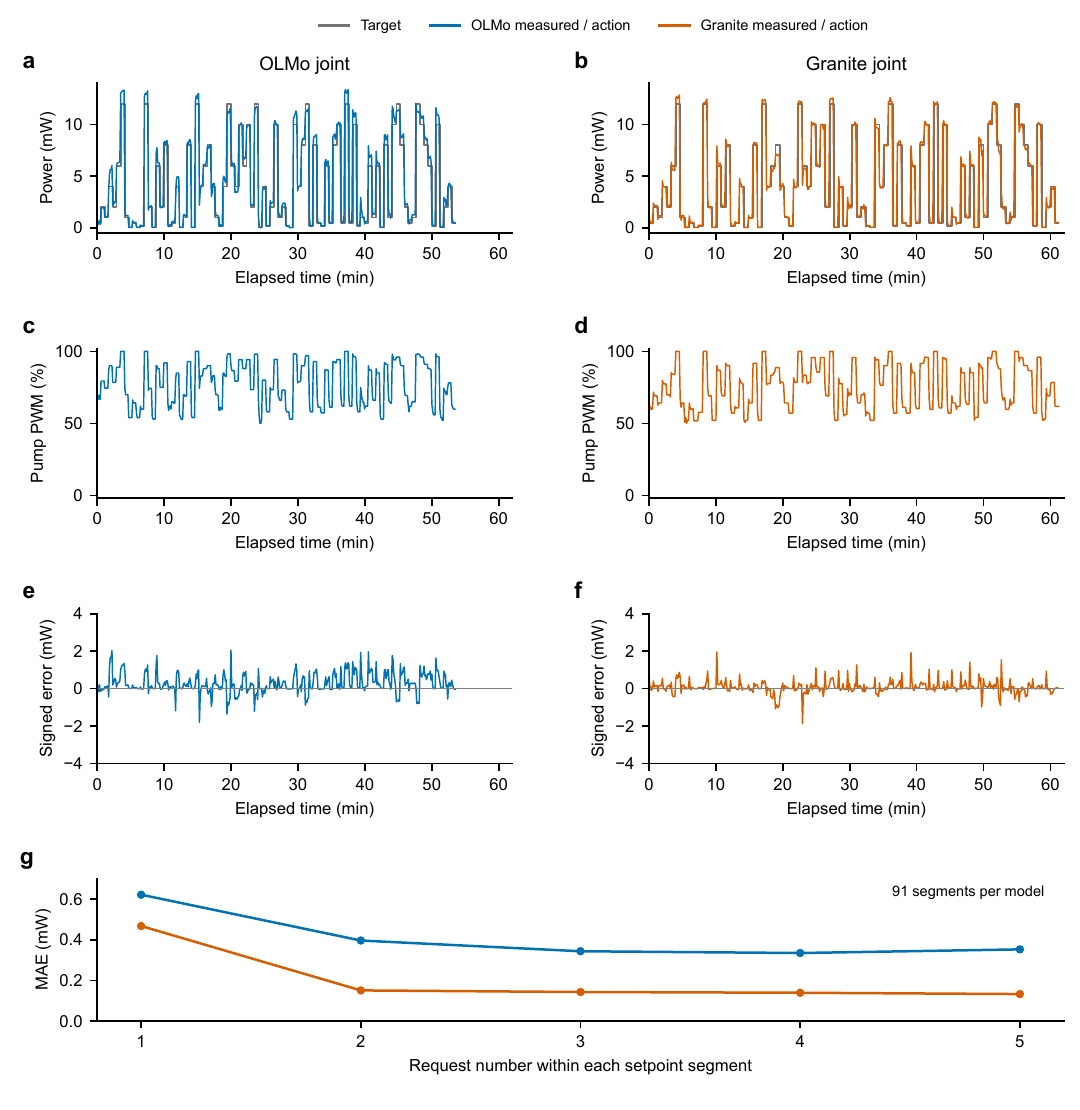}
\caption{\textbf{Continuous generator operation with two fixed joint models.} Left and right columns show OLMo and Granite, respectively, for all 455 requests in each recorded run. \textbf{a,b}, Requested and measured electrical power. \textbf{c,d}, Applied pump PWM. \textbf{e,f}, Signed measured-minus-target power error. \textbf{g}, Mean absolute error at each request position within a target segment, averaged over all 91 segments per model. Colors identify models; gray denotes target power in a,b. Each requested level is held for five model requests. Time is relative to the first outcome window, with identical axis ranges across the two models. Lines connect request-level measurements; the underlying sensor stream runs at 10~Hz. No model parameters are updated during either run.}
\label{fig:closed_loop}
\end{figure}

The joint OLMo model operated for 53.6~min at a median loop period of 7.0~s (Fig.~\ref{fig:closed_loop}a,c,e). Across all 455 requests, tracking MAE was 0.41~mW and RMSE was 0.60~mW. Median absolute relative error was 9.6\%, with 75\% of requests within 20\% of target. For targets of at least 0.5~mW, the corresponding values were 8.8\% and 79\%. MAE decreased from 0.62~mW on the first request of a segment to 0.35~mW on the fifth (Fig.~\ref{fig:closed_loop}g). Whenever the model changed the PWM between successive requests at the same target after an error above 0.1~mW, the change was in the error-reducing direction (75 of 75 changes). In the 61 segments in which the command changed at least once, MAE fell from 0.77~mW on the first request to 0.31~mW on the fifth. In the 30 segments in which the command was held, it remained between 0.33 and 0.44~mW. In 68\% of segments, error reached within 0.2~mW of target at least once; median time to first reach this tolerance among these segments was approximately 14~s. All commands were parsed and applied. The 10~PSI software pressure interlock was never triggered, and maximum recorded pressure was 3.21~PSI.

Physical operation also recurred with the larger Granite backbone (Fig.~\ref{fig:closed_loop}b,d,f), using the same protocol and target sequence. Its 455 requests spanned 61.4~min, with a median loop period of 8.1~s. Tracking MAE was 0.21~mW and RMSE was 0.36~mW. Median relative error was 4.6\%, with 83\% within 20\% of target; for targets of at least 0.5~mW, these values were 3.7\% and 92\%. First- and fifth-request MAE were 0.47~mW and 0.13~mW. All 74 such PWM changes were in the error-reducing direction. The physical-control result was therefore reproduced with a second model backbone.

These runs connect offline learning to sustained execution under real sensing and actuation. In the OLMo run, median within-window power variation was 0.12~mW and median actuator response to 90\% of the commanded change was 0.3~s. The loop used each new measurement to infer the next action throughout the recorded run, with no deployment-time parameter learning.

\section{Discussion}\label{sec:discussion}
This study demonstrates the joint acquisition of heterogeneous physical operating mappings within a single dense language model. Mappings with different governing physics, state descriptions and action spaces coexist with numerical errors matching those of separately trained models. Physical explanations are learned alongside executable commands, and continuous generator operation with OLMo and Granite demonstrates execution under measured feedback as the agent's own actions change its subsequent inputs. Together, these results connect accumulated operating records, shared model parameters and physical execution.

The shared operating intelligence connects requested power and available state information to numerical commands across distinct energy domains. The same model generates reactor reactivity and ramp duration, turbine fuel and feedwater flows, and generator pump PWM, together with physical explanations. The corresponding operating mappings span transient reactor kinetics, steady-state thermodynamics and hydraulic-electric conversion. Their joint acquisition brings heterogeneous physical operating knowledge into a common language interface and shared parameters without domain-specific modules.

The training formulation makes physical explanations an explicit learning target alongside executable actions. Operating records are augmented with explanations relating states, objectives and actions, and supervised fine-tuning trains the model to produce both the explanation and the numerical command. The resulting interface presents the requested objective, proposed action and stated physical justification together for operator review. Explanation generation is therefore part of the acquired operating capability.

Open weights provide complementary evidence about the internal organization of the acquired capabilities. The pretrained comparisons show that domain separation precedes operational training and persists as within-domain representations change. Objectives are accurately decodable across the evaluated models, while the probability assigned to an originally generated reactivity value responds differently to small reactor-target changes across backbones. Objective decodability and output-probability sensitivity therefore describe different aspects of the acquired mappings. These observations distinguish inherited domain structure, changes associated with operational training and differences in objective dependence across models.

In this formulation, operating records and simulators support capability development, while measured feedback supports successive application of the acquired mappings. Offline reinforcement learning, imitation learning and sim-to-real control provide approaches to learning from prior data and simulated experience \cite{levine2020offline,garcia2015comprehensive,brohan2023rt}. Simulation-trained controllers have demonstrated tokamak plasma control and champion-level drone racing \cite{degrave2022magnetic,kaufmann2023champion}. HydroGym supports the acquisition and evaluation of flow-control policies in simulation, including transfer from inexpensive surrogate environments to more demanding flow configurations \cite{lagemann2026hydrogym}. The present study complements operator-based virtual sensing, which recovers states that cannot be measured \cite{kobayashi2026virtual}. For the energy systems studied here, operating records train the numerical mappings and their accompanying explanations within shared parameters, and new generator measurements supply the state for each subsequent command.

Operating records connect physical knowledge to the learned actions. Physics-informed machine learning provides a broader framework for combining physical knowledge with data-driven modeling \cite{karniadakis2021physics}, with applications including transfer learning for predicting nuclear-reactor transients \cite{prantikos2023physics}. In the present study, forward simulators generate training examples and evaluate inferred actions, while measured generator records connect the learning process to the physical testbed. The language model learns inverse operating mappings from these examples without an explicit governing-equation residual in its training objective.

The results cover nominal operation in the supplied domains and within the tested envelopes. Reactor and turbine evaluation uses the same forward-model families as data generation, and hardware evidence comprises laboratory runs lasting approximately one hour. Transfer to additional physical systems, fault handling, emergency response, distribution shifts and long-term reliability are subjects for further evaluation \cite{ashmore2021assuring}. The faithfulness of the generated explanations, their benefit to operators and the internal components that causally support action generation \cite{wang2022interpretabilitywildcircuitindirect} also remain to be examined.

These results establish a shared language-based approach to acquiring and executing operating capabilities across distinct energy domains. Operating records and simulators support capability acquisition before deployment, while measured feedback conditions successive actions with fixed model parameters. The generator runs demonstrate sustained physical execution with two model backbones. Together, these findings provide a basis for organizing energy-system operating knowledge within a common model and invoking the corresponding capabilities through human-readable states and objectives.

\section{Methods}\label{sec:methods}

\subsection{Problem formulation}\label{subsec:formulation}

Each physical system was formulated through direct inverse action inference. Let $\mathcal{S}_d\colon\mathcal{Z}_d\times\mathcal{X}_d\rightarrow\mathcal{Y}_d$ denote the forward response of system $d$, where $\mathbf{z}\in\mathcal{Z}_d$ is the observed state, $\mathcal{X}_d$ is the defined action space and $y^*\in\mathcal{Y}_d$ is the operating objective. The objective is to identify an action that brings the system to the requested objective within its evaluated operating envelope,

\begin{equation}\label{eq:inverse}
    \mathbf{x}^*\in\operatorname*{arg\,min}_{\mathbf{x}\in\mathcal{X}_d}\left|\mathcal{S}_d(\mathbf{z},\mathbf{x})-y^*\right|.
\end{equation}

The inverse mapping was approximated by an autoregressive language model $f_\theta$ that, given a domain identifier, the observed state and the objective, generated a structured physical explanation followed by the action vector $\hat{\mathbf{x}}=f_\theta(d,\mathbf{z},y^*)$. Each direct-inference request comprised one generation episode without iterative optimization, simulator feedback or action revision, and the parameters $\theta$ were fixed after training. For the reactor task, the state was the fixed initial power and the objective was a target final power. For the turbine task, the objective was a target electrical power. For the physical testbed, the state comprised current PWM, flow, pressure, turbine voltage, generator current and electrical power and the objective was target electrical power. In closed-loop operation the same mapping was evaluated repeatedly, $\hat{\mathbf{x}}_t=f_\theta(d,\mathbf{z}_t,y^*_t)$, with $\mathbf{z}_{t+1}$ measured from the hardware after $\hat{\mathbf{x}}_t$ was applied.

\subsection{Nuclear-reactor simulator}\label{subsec:nuclear}
The nuclear benchmark represented a small modular reactor at an initial thermal power of 250~MW$_{\mathrm{th}}$ \cite{nrc2023nuscale}. Dynamics were simulated with PyRK, which solves point kinetics coupled to a lumped thermal-hydraulic feedback model \cite{huff2015pyrk}. The neutron population evolved according to

\begin{equation}\label{eq:point_kinetics}
    \frac{dn}{dt}=\frac{\rho(t)-\beta}{\Lambda}n(t)+\sum_{i=1}^{6}\lambda_i C_i(t),
\end{equation}

\noindent with total reactivity $\rho(t)$, delayed-neutron fraction $\beta$, prompt-neutron generation time $\Lambda$ and six precursor groups $C_i$. Fuel and coolant feedback coefficients were $\alpha_f=-3.0\times10^{-5}$ and $\alpha_c=-20.0\times10^{-5}~\Delta k$~K$^{-1}$. Each trajectory covered 300~s at 1.0~s resolution. Reactivity insertion began at 100~s as a linear ramp with total reactivity $\rho_{\mathrm{total}}\in[-0.020,-0.0001]~\Delta k$ and duration $\Delta t_{\mathrm{ramp}}\in[3.0,40.0]$~s. Evaluation targets were restricted to 20--240~MW.

\subsection{Rankine-cycle turbine simulator}\label{subsec:boiler}
The turbine benchmark used a steady-state implementation of the \text{\AA}str\"{o}m--Bell 160~MW drum-boiler system \cite{bell2000drum}, representing natural-gas combustion, heat transfer to feedwater, steam-state determination, turbine expansion and electrical conversion. For fuel and feedwater mass-flow rates $\dot{m}_f$ and $\dot{m}_w$, steam-outlet enthalpy was

\begin{equation}\label{eq:boiler_energy}
    h_{\mathrm{out}}=h_{\mathrm{in}}+\frac{\eta_{\mathrm{boiler}}\dot{m}_f\mathrm{LHV}_{\mathrm{CH_4}}}{\dot{m}_w},
\end{equation}

\noindent with $\eta_{\mathrm{boiler}}=0.90$, LHV $=50{,}000$~kJ~kg$^{-1}$, feedwater at 250~$^\circ$C and steam at 130~bar. Steam properties followed IAPWS-IF97 through pyXSteam \cite{pyxsteam,wagner2008international}. Steam was expanded to 0.05~bar with $\eta_{\mathrm{turb}}=0.85$ and $\eta_{\mathrm{gen}}=0.98$. The simulator is defined for $h_{\mathrm{out}}\le3{,}900$~kJ~kg$^{-1}$ and returns no output above this limit. Fuel flow was sampled from 2.7 to 10.0~kg~s$^{-1}$ and feedwater flow from 42.0 to 155.0~kg~s$^{-1}$. Together with the enthalpy limit, the feedwater bound caps achievable power at 224.7~MW, and the set of admissible actions vanishes as this cap is approached. The admissible feedwater interval is 51.5~kg~s$^{-1}$ wide at 150~MW, 3.6~kg~s$^{-1}$ at 219.5~MW and 0.2~kg~s$^{-1}$ at 224.5~MW. Evaluation targets were therefore restricted to the range from 60 to 218.4~MW.

\subsection{Physical hydraulic-electric generation testbed}\label{subsec:testbed}
A recirculating hydraulic loop comprised a reservoir, DC pump, YF-S201 flow sensor, pressure tap, fully open ball valve and micro-turbine coupled to a DC generator. Pump PWM was the only controlled actuator. An INA219 measured generator voltage and current across a fixed 47~$\Omega$, 5~W load; power was calculated from voltage and current. A Raspberry Pi read flow pulses and an ADS1115 pressure channel and supplied PWM through a BTS7960 pump driver. Supplementary Fig.~2 shows the hydraulic and electrical connections. The pump was a 12~V DC diaphragm unit with supplier-listed ratings of 4~L~min$^{-1}$ and 80~PSI (Amazon ASIN B01N75ZIXF). The turbine-generator was a Beduan G1/2-inch unit sold as the 12~V variant (ASIN B07PHR4K9Z). Flow was measured using a Taidacent YF-S201 sensor (ASIN B0813KX62M), and pressure using a 0--100~PSI transducer with a 1/8-inch NPT connection (ASIN B0CRQTNY9F). Sensor readings used nominal conversion factors without independent calibration against reference instruments. At the start of each closed-loop run, pressure and current offsets were estimated over 2~s with the pump off and subtracted from subsequent readings. Pump PWM was generated at 1~kHz, and flow was calculated from pulse frequency using a nominal conversion factor of 7.5~Hz per L~min$^{-1}$. The pressure signal was read through a 10~k$\Omega$/22~k$\Omega$ voltage divider and converted using a nominal 0.5--4.5~V span corresponding to 0--100~PSI. Generator current was calculated from the INA219 shunt-voltage measurement using a nominal 0.1~$\Omega$ shunt resistance.

Sensors streamed continuously at 10~Hz. Each prompt used the latest 1~s window. A generated PWM command was applied immediately and the outcome was the 1~s window ending 2~s later. A software pressure interlock aborted operation at 10~PSI and forced pump output low on exit. In the reported joint OLMo run, mean model latency was 4.86~s (maximum 6.16~s) and median loop period was 7.0~s; Granite mean latency was 5.88~s with an 8.1~s median period. Hardware inference ran locally on an Apple M5 Max in fp16.

Targets were 0.05, 0.2, 0.5, 1, 2, 4, 6, 8, 10 and 12~mW. An Eulerian traversal of ordered distinct target pairs (seed 42), with an initial segment, gave 91 segments of five requests each. The first request of each segment defines single-request performance; all 455 define closed-loop performance. The requested target sequence is identical across models, and the measured starting states evolve under each model's preceding actions. Hardware evaluation had no held-out test set. The reported time to tolerance is the interval from the first state window to the first outcome within 0.2~mW of target in a segment; its median is computed over segments that reach this tolerance. It does not require subsequent requests to remain within tolerance.

\subsection{Data generation and explanation augmentation}\label{subsec:data}
For each simulated system, sampled actions were executed in the forward simulator and finite, successful records retained until 10,000 examples were collected. Each corpus used 8,000 training and 2,000 validation examples with seed 42 and stratification over ten objective bins. The hardware corpus contained 10,000 measured operations spanning pump PWM 0--100\%, flow up to 4.6~L~min$^{-1}$ and power up to approximately 14~mW. Its recorded split used 8,000 training and 2,000 validation records with source-aware, power-stratified groups of 20 consecutive operations and no group overlap. This training/validation split is distinct from the online hardware evaluation sequence.

Numerical examples were augmented with structured explanations relating the state, objective and action to the system physics \cite{hsieh2023distilling}. Explanations for all three domains were generated using Gemini 3 Flash through the API model identifier \texttt{gemini-3-flash-preview}. Explanations were spot-checked manually. For the generator corpus, automated checks enforced the required output sections and agreement between generated PWM values and recorded action labels. Explanations are synthetic training annotations. Joint training combined the three corpora, giving 24,000 training and 6,000 validation examples.

\subsection{Language-model backbones}\label{subsec:model}
The primary model was \texttt{allenai/Olmo-3-1025-7B}, with approximately 7.3 billion parameters, 32 decoder layers and hidden width 4,096 \cite{groeneveld2024olmo}. The second was \texttt{ibm-granite/granite-4.2-8b}, with approximately 8.8 billion parameters and 40 decoder layers \cite{granite2026}. Training used bfloat16 precision and gradient checkpointing. The downloaded OLMo and Granite model revisions were \texttt{a81bae42db3975be1671e27b9c9a56da1a9f980f} and \texttt{41a8a2d41c54ef4a71741b3e62604f4caaec9295}, respectively.

\subsection{Training procedure}\label{subsec:training}
Each example was formatted as one sequence containing a domain identifier, state and objective in the input portion and an explanation followed by the action variables in the output portion. Continued pre-training used packed 2{,}048-token blocks, 2 epochs, effective batch size 16, learning rate $2\times10^{-5}$ with AdamW, weight decay 0.01, cosine decay and 5\% warm-up. Supervised fine-tuning masked the identifier, state and objective tokens with the ignore index and computed loss only over the explanation and action outputs. Numerical tokens received a weight of $\lambda=2$ according to

\begin{equation}\label{eq:double_loss}
    \mathcal{L}_{\mathrm{SFT}}=\frac{\sum_{t\in\mathcal{T}_{\mathrm{out}}}w_t\ell_t}{\sum_{t\in\mathcal{T}_{\mathrm{out}}}w_t},
    \qquad
    w_t=
    \begin{cases}
        \lambda, & y_t\in\mathcal{N},\\
        1, & y_t\notin\mathcal{N},
    \end{cases}
\end{equation}

\noindent where $\ell_t=-\log p_\theta(y_t\mid y_{<t})$ and $\mathcal{N}$ is the set of numerical tokens. Supervised fine-tuning used 3 epochs, effective batch size 16, learning rate $1\times10^{-5}$, seed 42 and lowest-validation-loss checkpoint selection. The joint model used the balanced three-system corpus of 24,000 training and 6,000 validation examples with domain-and-bin stratification, and no domain-specific adapter, expert module, output head or parameter partition. Training ran on NCSA DeltaAI using one NVIDIA GH200 GPU per job.

\subsection{Inference and evaluation}\label{subsec:inference}
Reactor evaluation used 45 equal-width bins over 20--240~MW with 100 targets per bin; turbine evaluation used 24 bins from 60 to 218.4~MW with 50 targets per bin. The same requested targets were used for the domain-specific and joint models. Greedy decoding produced one output per request. Numerical action variables were extracted by regular expressions. Reactor actions were executed over the full 300~s transient with fuel and coolant temperature checks of 1,200 and 345~$^\circ$C, respectively. A sweep that also covered targets from 218.4 to 225~MW is not part of the evaluation. In that range, 39 joint and 35 domain-specific commands out of 50 exceeded the enthalpy limit, by less than 1.5\%, and the simulator returned no output. This is a limit of the simulator's property range and not a failure condition of the plant. For the operating maps in Fig.~\ref{fig:regimes}b, reactor iso-power contours were obtained by smoothed thin-plate-spline interpolation of final power over reactivity and ramp duration for the 10,000 executed PyRK records, restricted to their convex hull. Turbine contours were evaluated on a regular grid with the steady-state cycle model described above, which reproduces the recorded power of all 1,200 turbine executions to within $10^{-12}$~MW. The valid turbine region corresponds to steam quality of at least 0.95 and outlet enthalpy of at most 3,900~kJ~kg$^{-1}$.

MAE, RMSE, threshold accuracies and cumulative accuracy curves use all requests in the evaluation set. Hardware MAE compares measured power with the requested target; absolute relative error divides that error by the target. Both all-target and at-least-0.5~mW relative errors are reported for closed-loop runs. All reported values are descriptive statistics of predefined sweeps and recorded hardware sequences.

\subsection{Internal-representation analysis}\label{subsec:methods_interpretability}
Prompt representations were mean-pooled over input tokens at the embedding output and each decoder output. We selected 450 equally spaced prompts per domain from ordered reactor and turbine sweeps and the corresponding model's recorded hardware run. All supplied prompt tokens, including the identifier, were retained. PCA reduced representations to at most 50 components before silhouette calculation with the three domain labels \cite{rousseeuw1987silhouettes}. The pretrained backbone was encoded on the same prompts as its adapted counterpart. Linear centered kernel alignment compared pretrained and adapted representations within each domain at corresponding depths \cite{pmlr-v97-kornblith19a}. These analyses quantify domain separation and within-domain representational changes under the supplied prompts.

For the layer-16 read-out in Fig.~\ref{fig:regimes}a, we used the joint OLMo model's prompt-token-averaged representations at the output of decoder layer 16 for the same 450 prompts per domain. We selected 32 channels at evenly spaced indices $\operatorname{round}(4095j/31)$, $j=0,\ldots,31$, without data-dependent selection or reordering. Each channel was standardized using its mean and population standard deviation across all 1,350 prompts before averaging within each domain. The color scale was fixed from $-2$ to $+2$ standard deviations. The read-out summarizes standardized channel responses to prompts from each domain.

\subsection{Linear probes and objective perturbations}\label{subsec:probes}
We analyzed the query-position hidden state immediately before the first action value in the reactor and turbine sweeps (4,500 and 1,250 requests, respectively) for joint OLMo CPT followed by SFT, OLMo SFT-only and Granite CPT followed by SFT. Domain identity was decoded by class-balanced logistic regression, and objective power by ridge regression with regularization parameter 1. A grouped 80/20 train/test split (seed 42) kept identical domain and target values rounded to 0.01~MW in the same partition. Standardization and PCA to at most 50 components were fitted only on the training partition. Objective probes were fitted separately within each domain and evaluated on 893 reactor and 254 turbine cases.

For reactor prompts, we masked the objective or shifted it by $-1\%$ or $+1\%$ and measured the change in log-probability assigned to the original reactivity value. Supplementary Fig.~3c reports medians and interquartile ranges over valid cases. This measures the sensitivity of the original output probability to the objective.

\bmhead{Data availability}
The simulator-generated and testbed datasets supporting this study, together with the processed data underlying all figures, are available at \url{https://github.com/sixticket/text-based-agent}.

\bmhead{Code availability}
Source code for data generation, augmentation, training, inference, simulator and testbed evaluation, closed-loop control, representation extraction and figure generation is available at \url{https://github.com/sixticket/text-based-agent}, including environment specifications, model identifiers and revisions, random seeds, prompt templates and evaluation target files.

\bmhead{Acknowledgements}
The authors acknowledge the National Center for Supercomputing Applications (NCSA) at the University of Illinois Urbana-Champaign for providing the DeltaAI supercomputing resources that contributed to this work.

\bmhead{Funding}
Authors acknowledge the support from the DOE Distinguished Early Career Award (DOE DE-NE009635).

\bmhead{Author Contributions}
Y.P.L. conceived the study and developed the learning regime and methodology; designed the text-based interface; generated the simulation and augmented datasets; developed and trained the models; built and operated the physical testbed; performed all simulation, closed-loop and representation-analysis experiments; analyzed the results; and wrote the original manuscript. S.R., K.K. and S.C. reviewed and edited the manuscript. S.B.A. provided supervision and computing resources and reviewed and edited the manuscript. All authors read and approved the final manuscript.

\bmhead{Competing Interests}
The authors declare no competing interests.

\bmhead{Supplementary Information}
Supplementary information is available for this paper.

\bibliography{sn-bibliography}

\clearpage
\section*{Supplementary Information}

\begin{figure}[H]
\centering
\includegraphics[width=\textwidth]{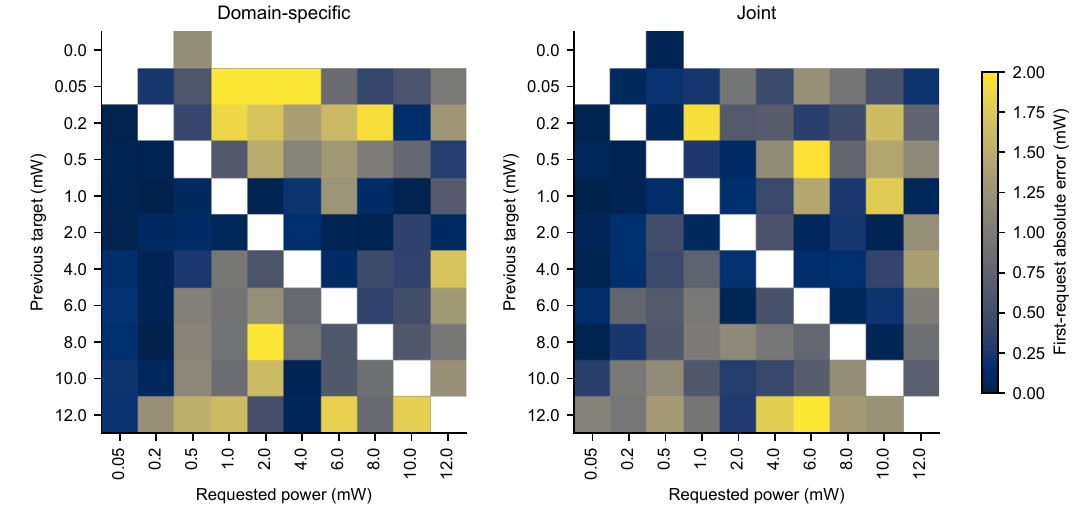}
\caption*{\textbf{Supplementary Fig. 1. First-request generator error by setpoint transition.} Left, domain-specific OLMo; right, joint OLMo. Each colored cell shows absolute measured-minus-target power error for the first request in a recorded setpoint segment. Rows give the preceding requested power and columns the new requested power; the initial segment starts from pump-off. White cells were not requested. The color scale is shared and saturates at 2~mW. These online measurements are the 91 single requests per model used in Fig.~\ref{fig:three_systems}g--i.}
\end{figure}

\clearpage
\begin{figure}[H]
\centering
\includegraphics[width=\textwidth,trim=0 353 0 0,clip]{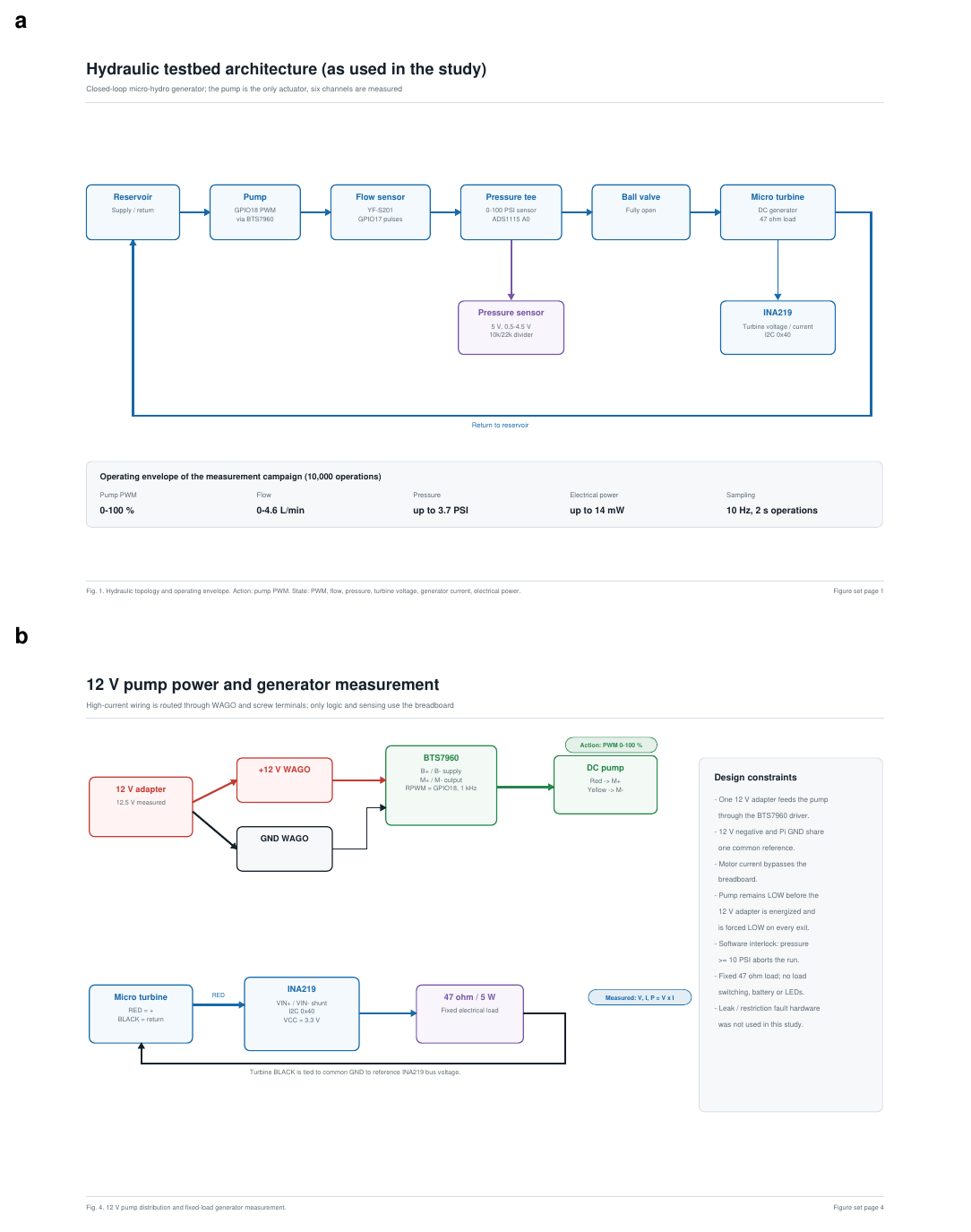}\\[4pt]
\includegraphics[width=\textwidth,trim=0 24 0 326,clip]{Figure/edfig2_hardware.pdf}
\caption*{\textbf{Supplementary Fig. 2. Hydraulic and electrical configuration of the generator testbed.} \textbf{a}, Recirculating hydraulic circuit. Pump PWM controls flow through a micro-turbine coupled to a DC generator. \textbf{b}, Power and generator circuit, including the BTS7960 pump driver and INA219 measurement across a fixed 47~$\Omega$ load. The pressure interlock is implemented in software. The drawings describe the configuration used for the reported experiments.}
\end{figure}

\clearpage
\begin{figure}[H]
\centering
\includegraphics[width=\textwidth]{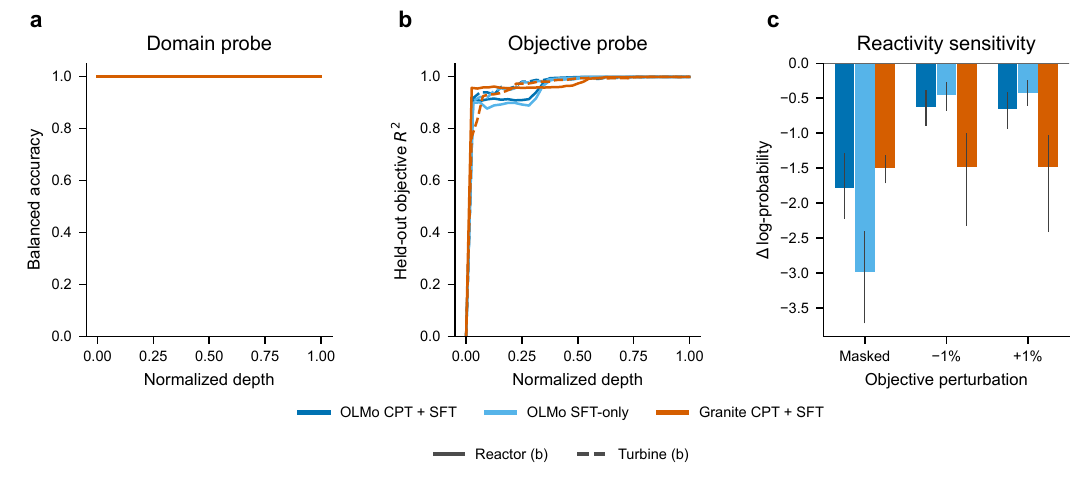}
\caption*{\textbf{Supplementary Fig. 3. Objective decodability and reactivity sensitivity.} \textbf{a}, Balanced accuracy of the linear domain probe at the query preceding the first action value. All three curves coincide at 1.00. \textbf{b}, Held-out objective-power $R^2$ for reactor (solid) and turbine (dashed) ridge probes. Colors identify the three joint models. \textbf{c}, Change in log-probability of the original reactor reactivity when the objective is masked or shifted by $-1\%$ or $+1\%$. Bars show medians and error bars the interquartile range across 4,500 cases per model. Probe preprocessing is fitted on the training partition only.}
\end{figure}

\end{document}